\documentclass[letterpaper, 10 pt, conference]{ieeeconf}  %

\IEEEoverridecommandlockouts                              %

\newcommand{\dGLIdiff}{$\text{dGLI Disk}_{\text{diff}}$}
\newcommand{\dGLIdiffsimp}{$\Delta\text{dGLI Disk}$}

\usepackage{xcolor} %

\usepackage{balance} %
\usepackage{hyperref} %
\usepackage{amsmath}
\usepackage{amssymb}

\usepackage{caption} %
\usepackage{graphicx}
\usepackage{subcaption}
\usepackage{cuted}
\usepackage{adjustbox}

\usepackage{booktabs}
\usepackage{multirow}

\usepackage{cite}

\author{%
Jay Kamat, Júlia Borràs, Carme Torras
\thanks{The research has been partially supported by project PID2023-152259OB-I00 (CHLOE-MAP), MCIU/ AEI /10.13039/501100011033, project ROBassist (CSIC code 202450E060), and by ERDF, UE and project SoftEnable (HORIZON-CL4-2021-DIGITAL-EMERGING-01-101070600)}%
\thanks{All authors are with the Institut de Robòtica i Informàtica Industrial, CSIC-UPC}%
}
\title{\Huge A Geometric Shape- and Orientation-Agnostic Cloth State Representation} %
\title{\Huge CloSE: A Geometric Shape-Agnostic Cloth State Representation}
\begin{document}

\maketitle

\begin{abstract}
    Cloth manipulation is a difficult problem mainly because of the non-rigid nature of cloth, which makes a good representation of deformation essential. We present a new representation for the deformation-state of clothes.
    First, we propose the {\it dGLI disk representation} based on topological indices computed for edge segments of the cloth border that are arranged on a circular grid. The heat-map of the dGLI disk uncovers patterns that correspond to features of the cloth state that are consistent for different shapes, sizes or orientation of the cloth. We then abstract these important features from the dGLI disk into a circle, calling it the {\it Cloth StatE representation} (CloSE). This representation is compact, continuous, and general for different shapes. We show that this representation is able to accurately predict the fold locations for several simulation clothing datasets. Finally, we also show the strengths of this representation in two relevant applications: semantic labeling and high- and low-level planning. 
    The code and the dataset can be accessed from: \url{https://close-representation.github.io/}
\end{abstract}

\section{Introduction}   
Cloth manipulation is a challenging problem in robotics mainly because of the infinite-dimensional configuration space of the cloth, i.e. the space of possible cloth positions (or states) in space. %
Unlike rigid or even articulated objects, where one just needs the object pose and joint configurations, clothes can deform in multiple ways making it extremely difficult to have a simplified representation. In addition, the various shapes, sizes, and mechanical properties of clothes add to the difficulty. For this reason,  end-to-end learning-based methods struggle to learn to manipulate clothes even in simulation, because by sampling examples it is nearly impossible to explore all of the configuration space. 
Recognizing and interpreting their deformation state with such intricate shapes is challenging even in simulation.
Several reviews in literature point to the need of a simplified good representations that could pave the way for having more efficient learning methods \cite{yin2021modeling, sanchez2018robotic,zhu2022challenges}. %

\begin{figure}[!t]
    \centering
    \includegraphics[width=0.9\linewidth]{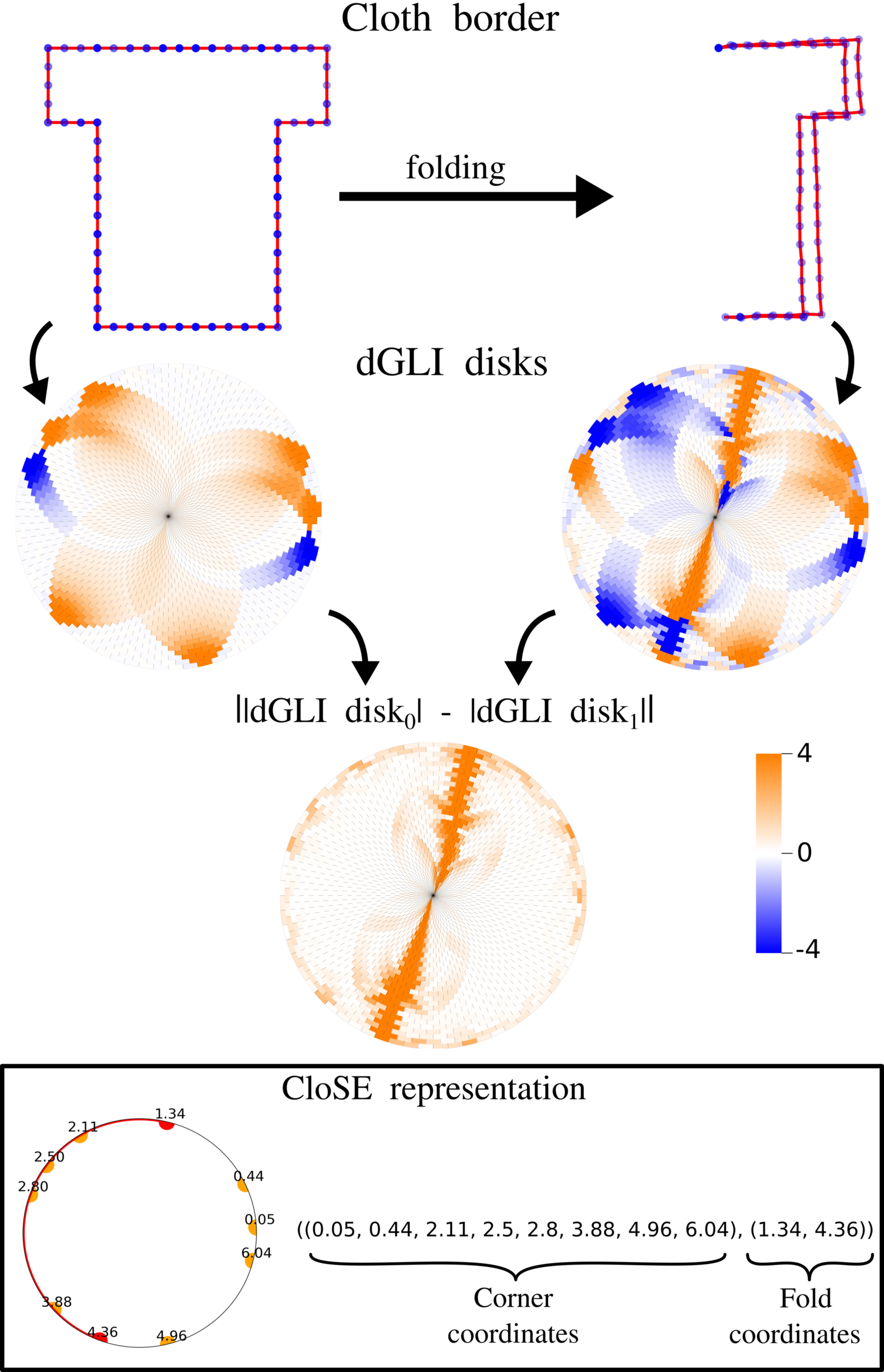}
    \caption{
    Example of CloSE derivation: From the border, we compute the dGLI disk where each petal corresponds to a corner. The difference between dGLI disks for the start and end frames allows to extract the fold information, so as to obtain the final CloSE descriptor that is shown graphically and numerically at the bottom of the figure.
    }
    \label{fig:PullFig-tshirt}
    \vspace{-15pt}
\end{figure}

In recent years, many datasets with both real and simulated clothes have emerged, as summarized in \cite{longhini2024unfolding}. Real images lack ground truth information on the configuration of cloth, and many reconstruction works \cite{bednarik2018learning,chi2021garmentnets} rely on realistic simulated images where the ground truth is the full mesh. But even then, there is no real understanding of the configuration of cloth other than comparing to predefined states or recognizing which pixels are sleeves or collars. In other words, there is no definitive solution to reason on the configuration space of cloth. In addition, partial solutions are costly and inefficient because robots could work with much less information than the full mesh. For instance, previous works have shown how recognizing salient features, corners or borders is enough for effective manipulation \cite{ramisa20163d, qian2020cloth}. Existing solutions to navigate the configuration space of cloth are based on RGB-D zenithal images of folded states \cite{lippi2022enabling, tanaka2018emd}, that effectively learn observed transitions between the silhouette of observed states.%

For folding tasks, cloth border is an informative indicator of the state of the cloth. Works such as \cite{Franco-dGLI} use the border of the cloth to characterize its state. In particular, they apply a derivative of the Gauss Linking Integral (GLI), a topological index that measures the linking number of curves, to the cloth border to define a representation that separates different cloth states into different clusters. Our paper takes inspiration from this work, but goes beyond by representing the GLI derivative matrix in a novel way that allows to define a very compact representation of the folding state of cloth. Our compact representation, the CloSE representation, is general to different cloth shapes and is also continuous and fully analytical.
    
    Our CloSE representation encodes:
    \begin{itemize}
        \item The shape of the cloth, i.e., the location of the corners of the silhouette of the flat unfolded state of the cloth
        \item The location of the folds and their orientations (which side folds up)
    \end{itemize}

    We propose a method departing from  the work of \cite{Franco-dGLI}, taking all edges along the border and calculating the dGLI. The first important novelty is that we arrange these values in a circular grid instead of a matrix, naming it the dGLI disk. %
    We then abstract out the corners and fold locations from the dGLI disk onto a circle to get the CloSE representation.
    Fig.~\ref{fig:PullFig-tshirt} shows the starting and final cloth borders of a folding action, their corresponding dGLI disks, the subtraction operation that allows to define the fold coordinates, and the CloSE representation obtained after abstracting out the features. 
    
    The contributions of this paper include

    \begin{enumerate} 
        \item The dGLI disk, a new arrangement for the dGLI coordinates in a disk instead of a matrix, which unravels a hidden structure that is
        general to any shape, size and pose of the cloth. 
        \item A novel representation, CloSE, that captures this structure in a compact and continuous way.
        \item Two applications of the CloSE representation: defining an automatic semantic description of the cloth state and planning manipulation sequences.
    \end{enumerate}

    \begin{figure}[t]
        \centering
        \begin{subfigure}[b]{0.47\linewidth} %
            \centering
            \includegraphics[width=\textwidth]{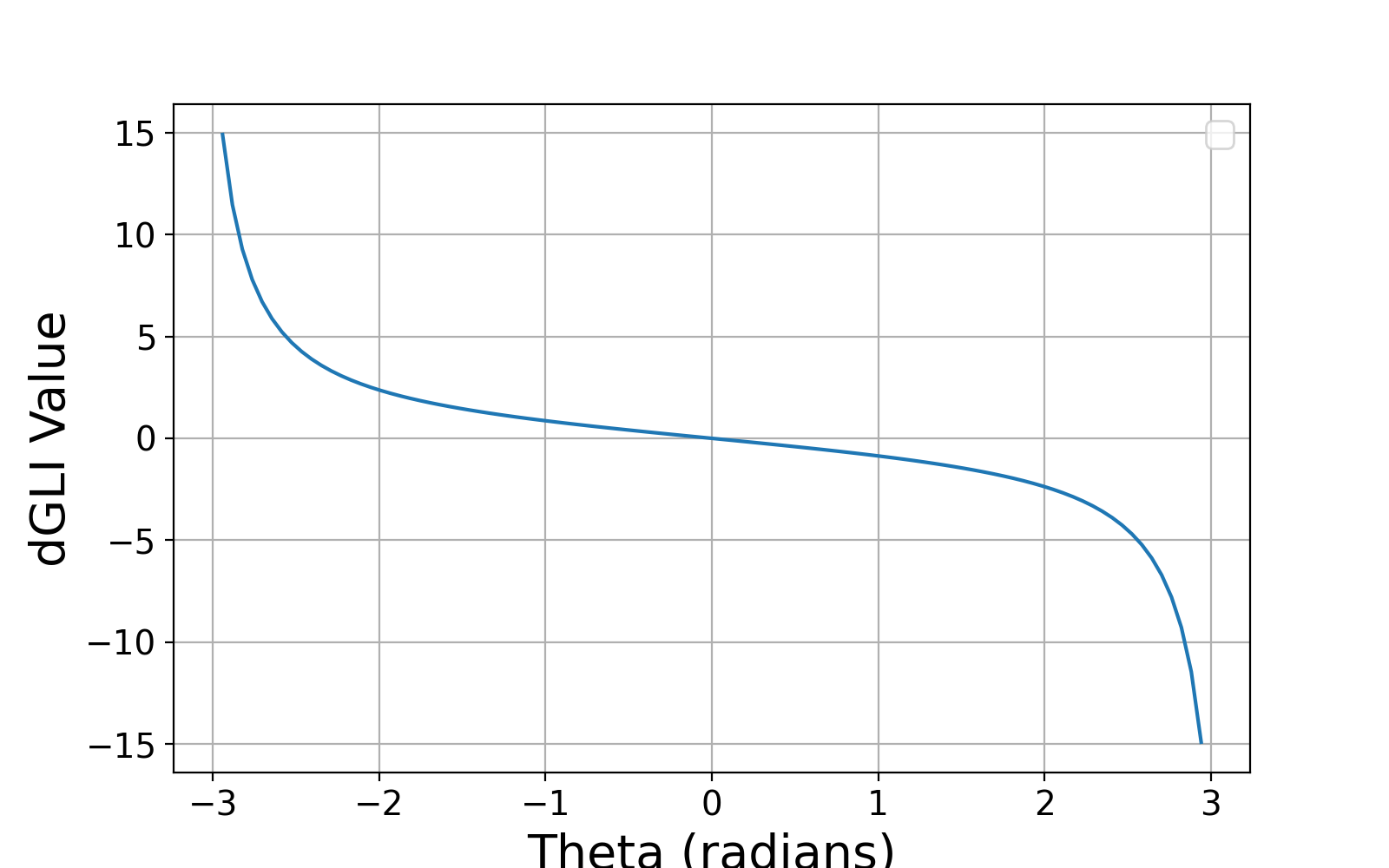} %
            \caption{Variation of dGLI with angle between segments}
            \label{fig:dglivsangle}
        \end{subfigure}
        \hfill %
        \begin{subfigure}[b]{0.47\linewidth} %
            \centering
            \includegraphics[width=\textwidth]{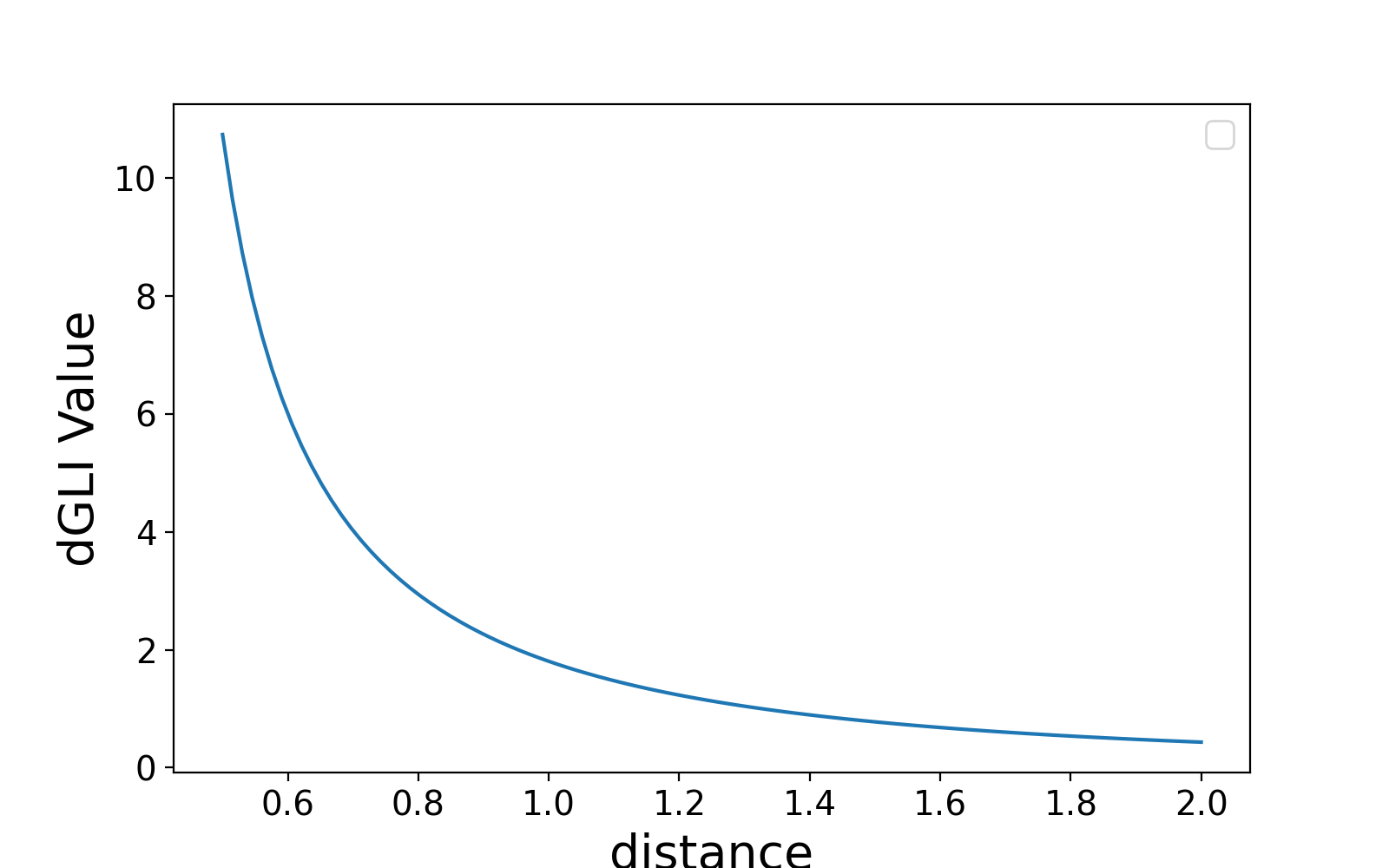} %
            \caption{Variation of dGLI with distance between segments}
            \label{fig:dgli vs distance}
        \end{subfigure}
        \caption{Behavior of dGLI in case of planar segments}
        \label{fig:dgli behavior}
        \vspace{-10pt}
    \end{figure}
    
\section{Related Work}

The configuration space of cloth is the space of possible cloth states in the Euclidean space. Different representations have been used in literature for robotic manipulation applications. Ranging from the full mesh of a simulated cloth \cite{huang2022mesh} to simplified geometric representations of the silhouettes \cite{miller2011parametrized,Doumanoglou2016,li2015folding}, none of these representations permit reasoning in this space in a general way, as they are highly dependent on the cloth shape. Later learning approaches relied on directly using the RGB or RGB-D as representation of the state of cloth \cite{hoque2020VisuoSpatial,seita2018deep,yan2021learning}, usually limited to zenithal views but still allowing to encode folding states.
One example is \cite{lippi2022enabling}, where a lower dimensional latent space of different observed states was encoded as a road-map to plan actions. Since these works map directly from the RGB image space to the latent space without any knowledge of the cloth (implicit or explicit), the latent space representation is not continuous, that is, the intermediate steps cannot be visualized. In addition, these data-based methods require a lot of training data for each cloth shape. Our proposed representation %
is general for any shape. 
     
The idea of using the Gauss Linking Integral (GLI) for robotics applications is not new. The GLI has been applied to representative curves of the workspace to guide path planning through holes \cite{ivan2013topology, zarubin2012hierarchical}, for guiding caging grasps in \cite{pokorny2013grasping, stork2013integrated, stork2013topology}, and for planning humanoid robot motions using the GLI to guide reinforcement learning \cite{yuan2019reinforcement}. The idea of applying the GLI to the curve that represents the border of the cloth was used in \cite{Franco-dGLI}. For planar curves the GLI is degenerated and clothes on a table are mostly planar, therefore, \cite{Franco-dGLI} proposed to apply a directional derivative of the GLI, leading to the dGLI coordinates, which enable distance reasoning in the C-space of cloth. This representation was used in \cite{Borras2023virtual} to learn semantic tags of a dataset of folding states of a squared piece of cloth. The dGLI coordinates rely on choosing a few segments in the border to calculate the dGLI, this makes them sensitive to noise in the chosen segments and may not be indicative of how the cloth behaves as a whole. Our work mitigates this by choosing all border edge segments for the dGLI calculation instead of a few. This reveals an underlying structure leading to the low-dimensional CloSE representation.

Our proposed method relies on knowing the curve of the silhouette border. For simulated clothes, we computed the border of the initial flat state since the full mesh is available. For real garments, recent advances have made this increasingly feasible: some works can reconstruct the full cloth mesh \cite{wang2024trtm}, while others can directly extract the border from images \cite{qian2020cloth,ren2023grasp}. However, these approaches are still in their early stages and their accuracy remains limited, which makes them less suitable for robust evaluation at this point. As these methods mature, we expect our representation to naturally extend to real-world cloth perception. However, we want to emphasize that our representation is useful in simulation to automatically label datasets and improve learning methods based on simulation data.

To the best of our knowledge, this is the first time a continuous representation of the folding state of cloth is proposed that is general across different cloth shapes.

\begin{figure}
    \centering
    \includegraphics[width=0.9\linewidth]{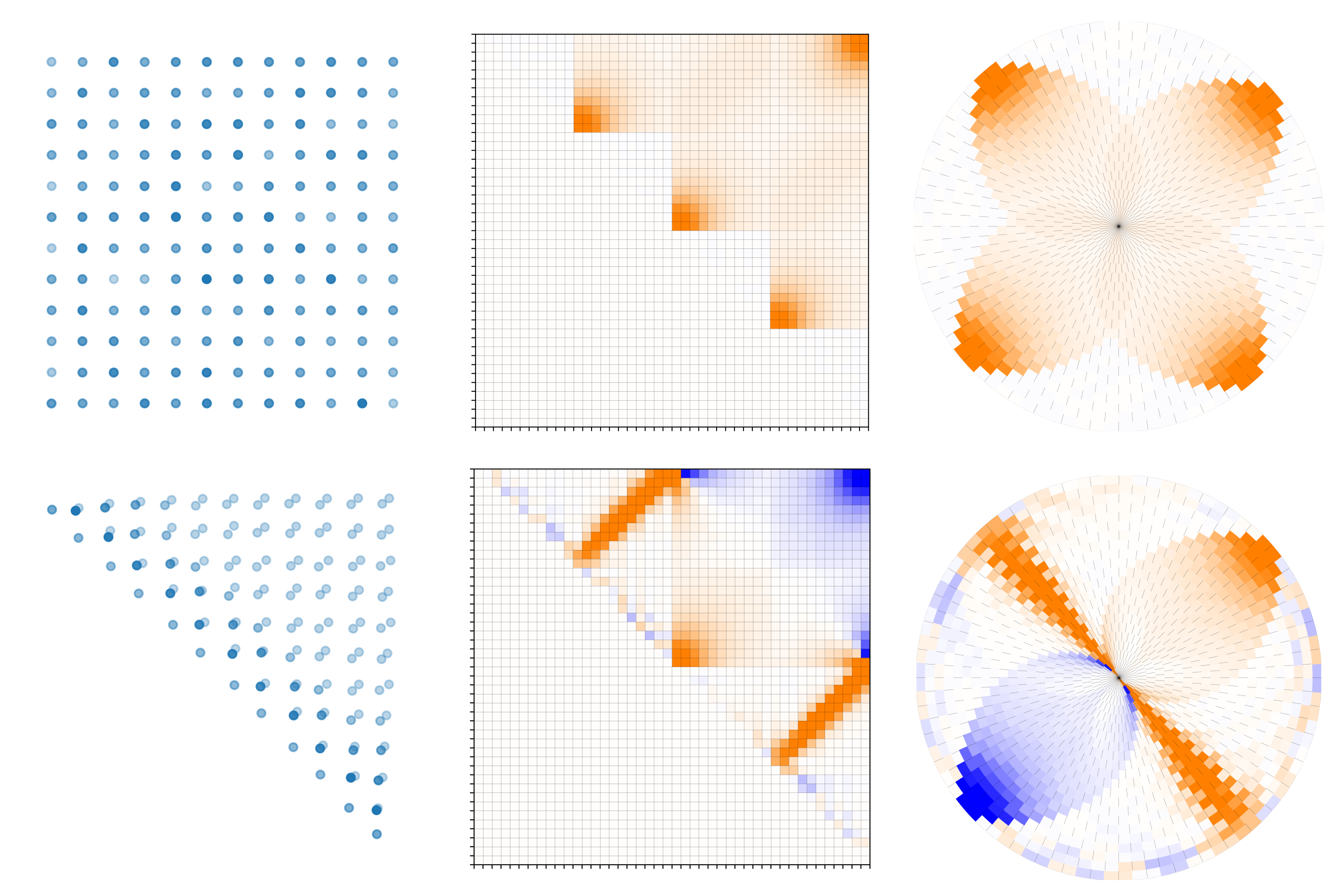} \\%
    a)\hspace{0.25\linewidth} b) \hspace{0.25\linewidth}   c)
    \caption{a) A square napkin mesh. b) The corresponding original dGLI matrix as in \cite{Franco-dGLI} and c) Our proposed arrangement in the dGLI disk}
    \label{fig:OldVsNew}
    \vspace{-10pt}
\end{figure}

\section{Background - \normalfont{d}GLI}
    The Gauss Linking Integral (GLI) Eq.\ref{eq: GLI} is a topological invariant that measures the linking number between two closed curves $\gamma_{1}$ and $\gamma_{2}$. 
    \begin{equation}\label{eq: GLI}
        \mathcal{G}(\gamma_{1} , \gamma_{2} ) = \frac{1}{4\pi}\int\int\frac{(\gamma_{2}-\gamma_{1})\cdot[\gamma_{2}' \times \gamma_{1}']}{||\gamma_{2}-\gamma_{1}||^3}
    \end{equation}
    
While originally defined for closed loops, the GLI can also be computed between two open curves to measure how much the two curves revolve around each other. The matrix formed by computing the GLI between different open segments has been used for various applications \cite{yuan2019reinforcement, ho2011thesis,stork2013topology}. For computational efficiency,~\cite{levitt1983protein} introduced an exact analytical formula to compute the GLI directly between straight line segments($\gamma_{AB}$ and $\gamma_{CD}$) (Eq. \ref{eq: ho-equation}). %
    
    \begin{align}\label{eq: ho-equation}
        \mathcal{G}(\gamma_{AB} , \gamma_{CD}) &= \arcsin(\vec{n_a}\cdot\vec{n_b}) + \arcsin(\vec{n_b}\cdot\vec{n_c}) \nonumber \\
        &\quad + \arcsin(\vec{n_c}\cdot\vec{n_d}) + \arcsin(\vec{n_d}\cdot\vec{n_a}) 
    \end{align}
    \begin{align*}
        \vec{n_a} &= \frac{\vec{AC}\times{\vec{AD}}}{||\vec{AC}\times{\vec{AD}}||}, \quad
        \vec{n_b} = \frac{\vec{BC}\times{\vec{BD}}}{||\vec{BC}\times{\vec{BD}}||}, \\
        \vec{n_c} &= \frac{\vec{CD}\times{\vec{CA}}}{||\vec{CD}\times{\vec{CA}}||}, \quad
        \vec{n_d} = \frac{\vec{DA}\times{\vec{DB}}}{||\vec{DA}\times{\vec{DB}}||}
    \end{align*}
    
Here, $A, B, C, D \in \mathbb{R}^{3}$, and $\vec{n_a}, \vec{n_b}, \vec{n_c}, \vec{n_d}$ represent the unit face normals of the tetrahedron formed by these four vertices. Geometrically, this formula evaluates the mutual twist of the two segments by calculating the signed solid angle, which corresponds to the spherical area of the quadrangle defined by these four normal vectors. A major limitation of using the measure GLI is that it vanishes when the segments are coplanar, which is a common occurrence in flat or folded fabric. To address this, \cite{Franco-dGLI} introduced the directional derivative of the GLI along the zenithal axis, the \textbf{dGLI} for cloth representation (Eq. \ref{eq: dgli}).
    
    \begin{equation}
        \text{dGLI}(\gamma_{AB}, \gamma_{CD}) = \frac{ \mathcal{G}(\gamma_{AB+\epsilon}, \gamma_{CD+\epsilon}) - \mathcal{G}(\gamma_{AB}, \gamma_{CD})}{\epsilon}
        \label{eq: dgli}
    \end{equation}
    
Here, $\epsilon$ is a small perturbation in the zenithal direction of the cloth made only to points $B$ and $D$ such that $B' = B + \epsilon\hat{k}$ and $D' = D + \epsilon\hat{k}$.
\footnote{For calculating dGLI, in this paper we use $\text{dGLI} = \frac{\mathcal{G}(x+2\epsilon) - \mathcal{G}(x+\epsilon)}{\epsilon}$\\ This does not change any of our analysis. We do this to handle consecutive edge segments.} 
In  \cite{Franco-dGLI} the \textit{dGLI coordinates} were defined as a matrix of all the crossed computations of the dGLI for a selected few edge segments on the cloth border. The dGLI coordinates could classify different cloth states into classes by distance. 

To understand the variations in dGLI for two segments, we numerically evaluate the dGLI for planar segments for varying angles and distances. The resulting graphs are shown in Fig. \ref{fig:dgli behavior}. Notice that the dGLI is exactly 0 when the lines are co-linear, and the sign flips based on the relative orientation of the two segments. The value of dGLI also decreases as the distance between the two segments increases. \looseness=-1

\section{Representation}

\begin{figure}
    \centering
    \includegraphics[width=\linewidth]{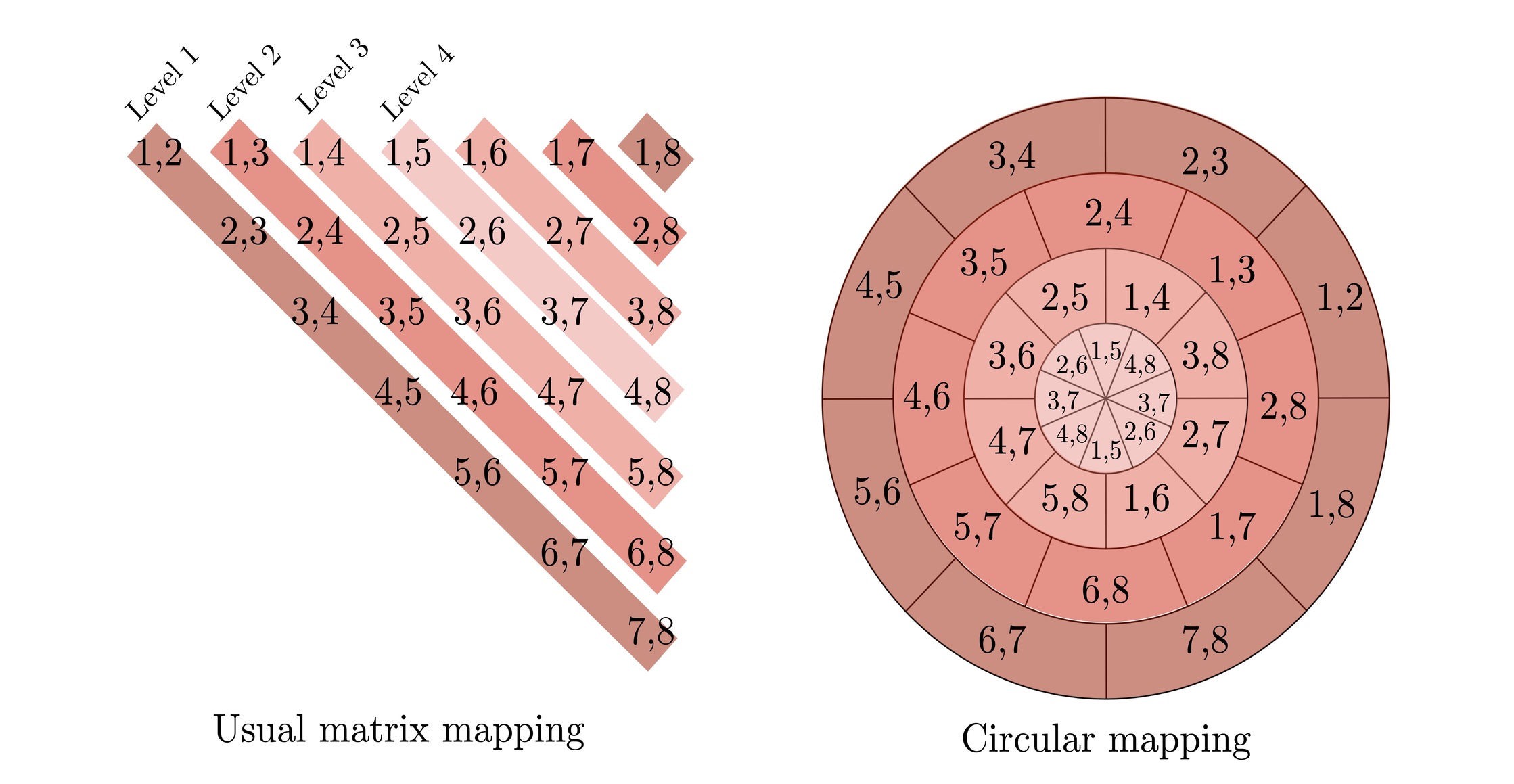}
    \caption{The Circular grid mapping in comparison to the semi-matrix mapping}
    \label{fig:CircularGridMapping}
    \vspace{-10pt}
\end{figure}

We propose a representation in two steps. First, the \textbf{dGLI disk}, a representation inspired by \cite{Franco-dGLI}, and  second, the \textbf{CloSE} representation, a compact and continuous representation that is abstracted out of the dGLI disk.

\subsection{The dGLI disk}
The dGLI disk is a novel circular arrangement for the dGLI matrix \cite{Franco-dGLI}. To illustrate, we take the example of a square napkin folded diagonally as shown in Fig. \ref{fig:OldVsNew}. 
Unlike in \cite{Franco-dGLI}, where the authors used only a few edge segments in the computation of the dGLI matrix, we use all the edge segments on the cloth border. This captures more information about the cloth. The matrix arrangement of the dGLI coordinates with all edge segments gives the heat-map in Fig. \ref{fig:OldVsNew}-b. Here, white represents zero values, and orange indicates positives and blue indicates negatives on a linear scale. In this form, we can already see interesting patterns. However, they are difficult to interpret as this matrix representation breaks the continuity of the cloth border at the first and the last indices.

To preserve the continuity of the cloth border in our representation, we re-map the matrix onto a disk, such that the dGLI value between immediate neighboring edge segments is mapped onto the outermost (first) layer of the disk. Interactions between second-neighbor segments are mapped to the second layer, and so on, moving inward. Consequently, the center of the disk represents the dGLI between segments on opposite sides of the cloth boundary. This radial logic is visually explained in Fig. \ref{fig:CircularGridMapping}. 
The new mapping (Fig. \ref{fig:OldVsNew}-c) displays easy-to-identify features. Notice a bright flower pattern in the unfolded $\text{dGLI}_{\text{disk}}$ indicating 4 corners. The folded $\text{dGLI}_{\text{disk}}$ adds an orange line between the two folded corners passing through the center, and the petals corresponding to the folded corners change colors. Our evaluations across 3 datasets consisting of pants, t-shirts, skirts and tops, shows that these patterns are consistent across all shapes.

\begin{figure}
    \centering
    \begin{subfigure}{0.49\linewidth}
        \centering
        \includegraphics[width=0.8\linewidth]{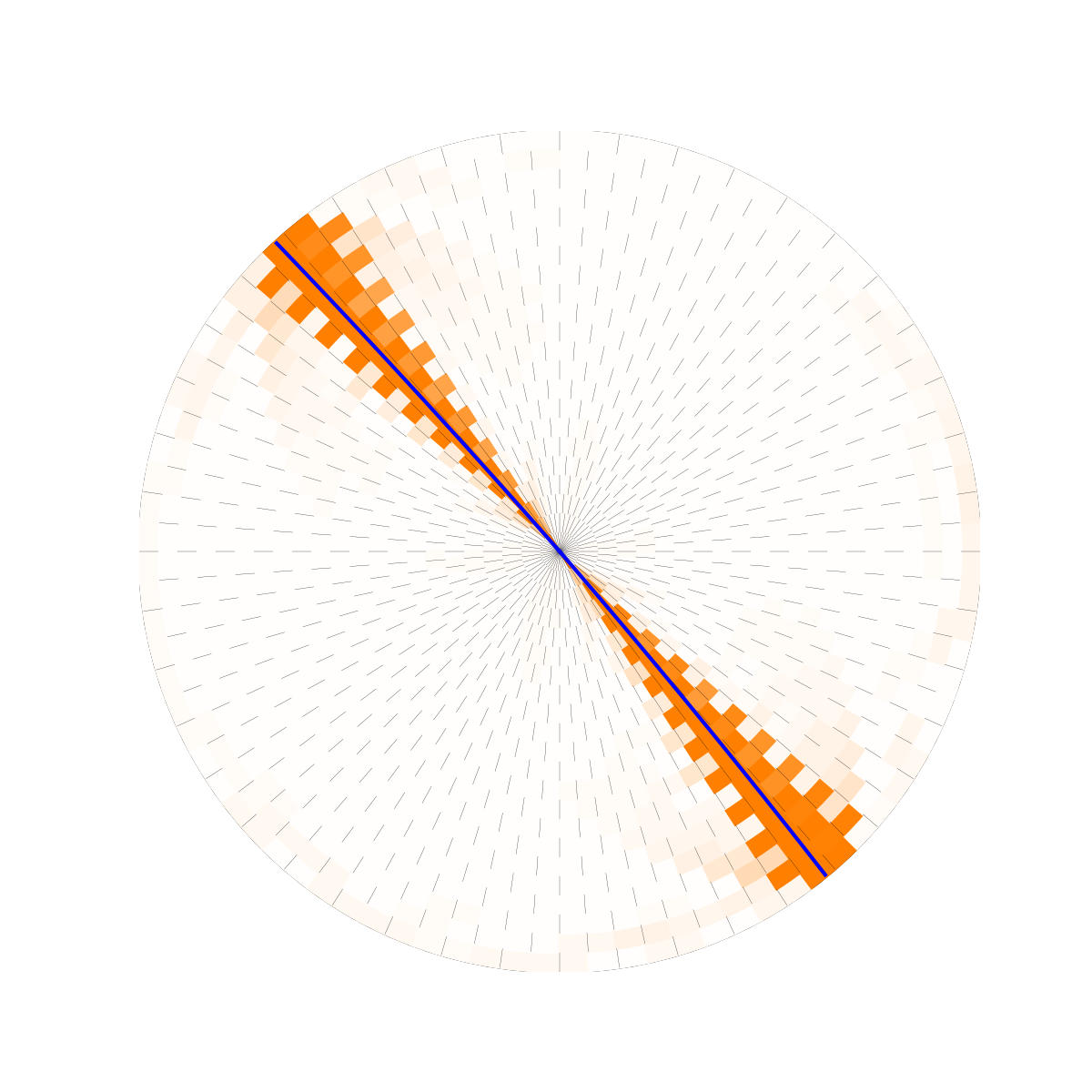}
    \end{subfigure}
    \hfill
    \begin{subfigure}{0.49\linewidth}
        \centering
        \includegraphics[width=0.8\linewidth]{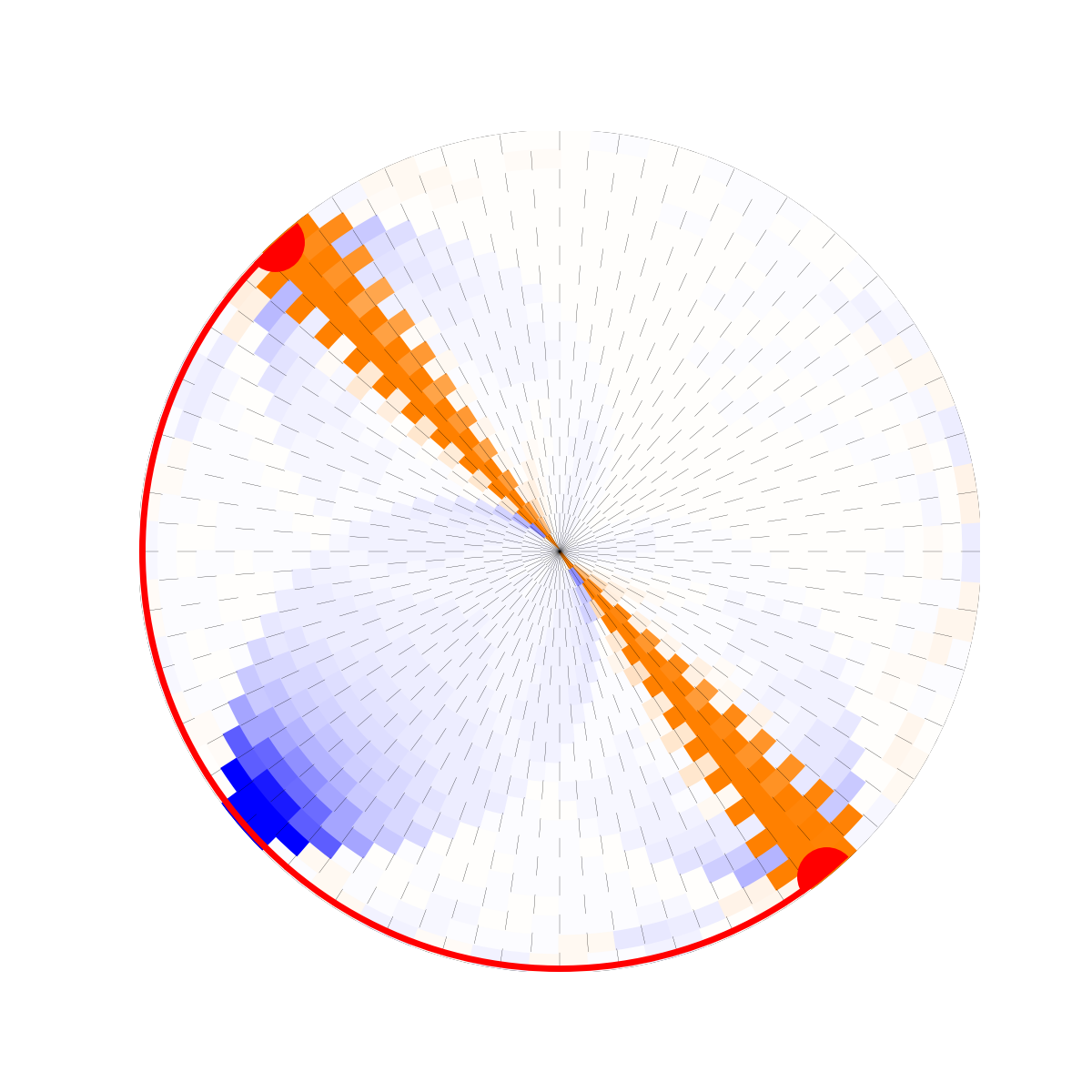}
    \end{subfigure}
    
    \caption{For the example in Fig. \ref{fig:OldVsNew}: (Left) The \dGLIdiff\ highlights the fold line. (Right) The \dGLIdiffsimp\ indicates the side folded.}
    \label{fig:2_dGLI_diffs}
    \vspace{-10pt}
\end{figure}

The pattern of the corners appears because higher dGLI values are observed when the edges are near and at an angle (refer Fig. \ref{fig:dgli behavior}). These dGLI values are positive or negative depending on their relative orientation. When folded, the angle reverses as seen from the zenithal direction, hence changing sign. In addition, a fold curve appears as the edge segments on the opposite sides of the fold come closer when the cloth is folded. 
Empirically, for single planar folds in our datasets, fold curves consistently pass through the center.
This fold curve can be easily isolated by taking the absolute difference between the dGLI disks, that is 
\begin{equation}
    \text{dGLI Disk}_\text{diff} 
        = \big|\,|\text{dGLI Disk}_\text{end}| - |\text{dGLI Disk}_\text{start}|\,\big|. \label{eq:dgli_diff_abs}
\end{equation}
Instead, the simple difference between the two dGLI disks helps us identifying the side folded, that is
\begin{equation}
    \Delta\text{dGLI Disk}
        = \text{dGLI Disk}_\text{end} - \text{dGLI Disk}_\text{start} \label{eq:dgli_diff_simp}
\end{equation}
Fig. \ref{fig:2_dGLI_diffs} shows the \dGLIdiff\ and the \dGLIdiffsimp\ for the example in Fig. \ref{fig:OldVsNew}.

\subsection{The Cloth StatE (CloSE) representation} \label{CloSE}

\begin{figure}[t]
    \centering
    \includegraphics[width=1\linewidth]{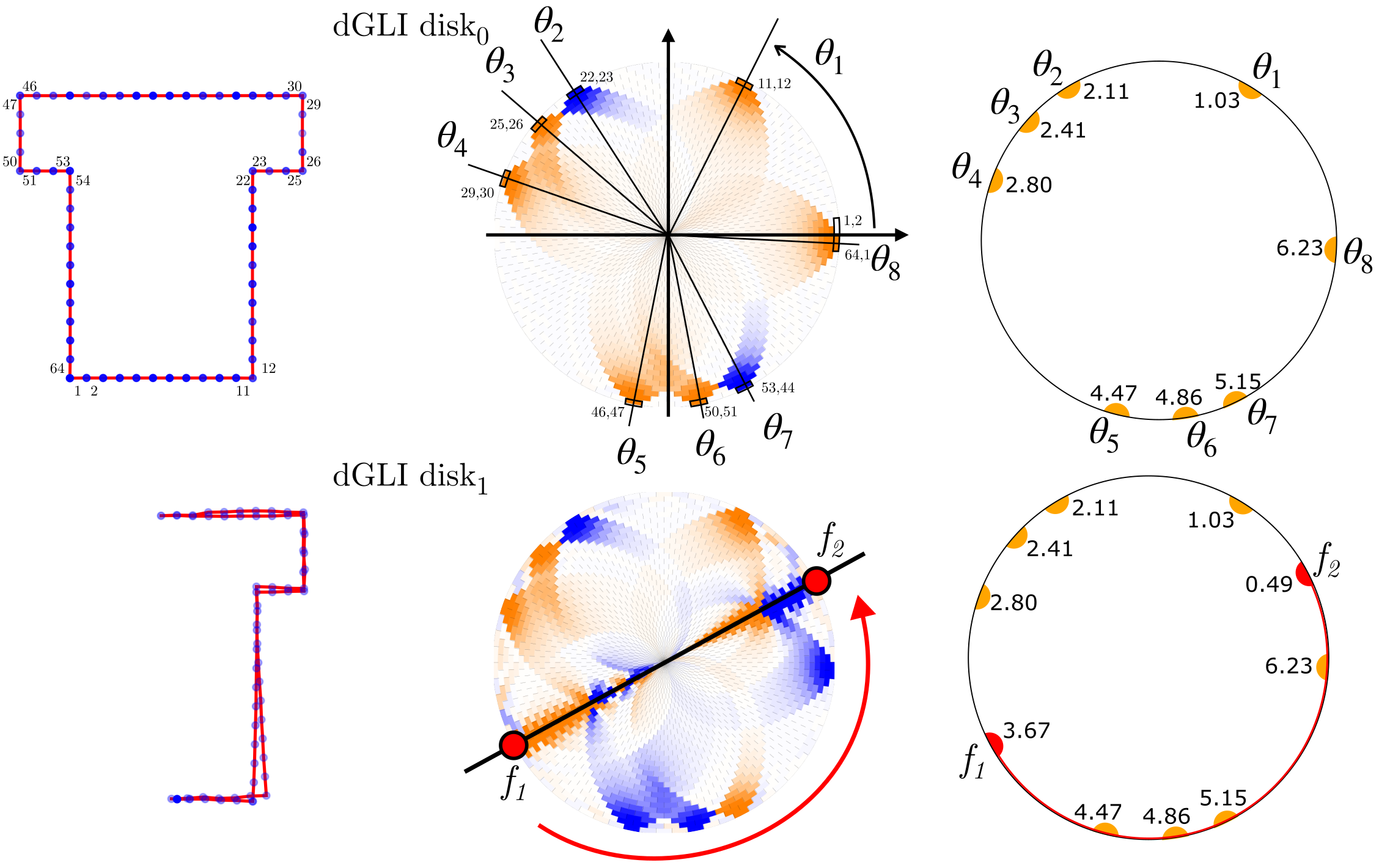}
    \caption{Computation of the CloSE representation from the dGLI disks. We abstract out the corner coordinates $\{\theta_1, \dots, \theta_8\}$ and the fold coordinates $(f_1,f_2)$ from the dGLI disk (middle) to the CLoSE representation (right).}
    \label{fig:closeComputation}
    \vspace{-10pt}
\end{figure}

The CloSE representation is a compact and continuous representation that abstracts out the corners and the fold curves from the dGLI disk into a circle\footnote{Intuitively, think of the CloSE representation as the cloth border morphed onto a unit circle}. The graphical description of how we computed is shown in Fig. \ref{fig:closeComputation}.

CloSE is a list of vectors, the first vector encodes the corner locations in radians, i.e., their location on the dGLI disk. The second represents the fold locations and orientation. Each fold is represented by the two end points of the fold line (in radians) while their order dictates the orientation of the fold. 
i.e., a fold $(f_1, f_2)$ will imply that the cloth border on the route from $f_1$ to $f_2$ in the anticlockwise sense is folded. 
This notation of fold definition ensures that all values of $(f_1, f_2)$ are valid folds, this allows for continuous linear interpolation between folds making the CloSE representation \textit{continuous}. Some examples of interpolations displaying the continuity in CloSE representation are shown in the accompanying video.

\subsubsection{Computation of the corner coordinates $\theta_1, \dots, \theta_n$}
Following the notation in Fig. \ref{fig:closeComputation}, to get the corners  we simply iterate through the first layer in the dGLI disk and note the location of the points that are above a certain threshold. Since the first layer only contains dGLI between immediate neighbors, one corner will only give one bright cell. 

When the mesh is wrinkly, it can give false corners that correspond to just wrinkles. However, we can still detect accurately the location of the folds and locate them with respect to the CloSE representation, as we will show in our results. The power of our method is that even if all the points of the border are considered corners, we can understand the fold state.

\subsubsection{Computation of the fold  coordinates \texorpdfstring{\normalfont$(f_1, f_2)$}{(f1, f2)}}

The curve(s) that we can see on the \dGLIdiff\ represent the fold line(s). We employ a curve fitting method to automatically approximate this fold curve (s). 

Given a \dGLIdiff, we arrange all its values in a vector 
$\{g_i\}$.
We fit a curve of the form $\theta = a_0+ a_1 r$, where $r$ and $\theta$ are radius and angle in polar coordinates. For simplicity, we only consider $r > 0$. So, two opposite curves represent one fold curve. 
To fit the curve, we use a gradient-based optimization algorithm \footnote{L-BFGS-B method in scipy.optimize.minimize from the SciPy library} to find the minimum of 
\begin{equation}\label{eq: loss}
    \text{loss} = \sum_i k_1\cdot(\theta_i - \hat{\theta}_i)^2 \cdot (p_1 \cdot \sqrt{g_i}) 
    - k_2 \cdot p_2 \cdot g_i
\end{equation}
 where $g_i$ is the $i$th value of the \dGLIdiff in the vector form $\{g_i\}$, $\theta_i$ is the angle coordinate corresponding to the value $g_i$ when arranged in the disk and $\hat{\theta}_i=a_0+ a_1 r_i$, being $r_i$ the radius coordinate to the same $g_i$ value in the disk. In addition, \looseness=-1
 \begin{align}
    p_1 &= \exp(\frac{-(\theta_i - \hat{\theta}_i)^2}{2\cdot\tau^2})\\
    p_2 &= \frac{1}{k_3+(\theta_i - \hat{\theta}_i)^2} 
\end{align}
with  $k_1, k_2, k_3, \tau \in \mathbb{R}$ being the parameters of the system.
We define 20 equidistant curves around the disk as initial points of our optimization, to ensure we converge to all the local minima. We then identify all the local minima with negative loss, each of them being a fold curve.

The loss function is made of two terms, the first term is positive: $k_1(\theta - \hat{\theta})^2 \cdot (p_1\sqrt{d\mathcal{G}_i})$, penalizes for having high error in curve fitting while also giving importance to the $g_i$ values nearby by weighting it with the term $p_1$. This term alone encourages local minimas that are far away from the points of interest, so, we add a second negative loss term, $k_2 \cdot p_2 \cdot g_i$, encouraging the curve to move near the points of interest. The only local minima that correspond to fold lines are the negative ones, where the second term is dominant. %
For each of the curves obtained $(a^i_0,a^i_1)$, the corresponding fold coordinate $f_i$ is obtained evaluating at $r=1$, that is, $f_i = a^i_0+a^i_1$. For single folds, we will obtain two fold coordinates. For multiple folds, preliminary results are shown in Section \ref{sec:multi-folds}.

\subsubsection{Ordering of $(f_1, f_2)$ to encode the fold orientation}
Corners affected by the fold change sign. Taking the absolute value of \dGLIdiffsimp, non-zero values are always on the same side of the fold curve, the side that is folded. We order the fold coordinates $(f_1, f_2)$ such that the folded corners lie between $f_1$ and $f_2$ while traversing in the anti-clockwise sense.  %

\begin{table}[tb] 
\centering
\caption{Evaluation results} \label{tab:eval}
\begin{tabular}{lcc}
\toprule
\textbf{Dataset} & \textbf{RMSE (Mean, Var)} & \textbf{Frechet (Mean, Var)} \\
\midrule
Clothilde & 0.05, 0.03 & 0.09, 0.05 \\
Napkin  & 0.07, 0.02 & 0.11, 0.04 \\
\midrule
\multicolumn{3}{l}{\textbf{VR\_folding (First Folds)}} \\
Tshirts & 0.13, 0.07 & 0.21, 0.08 \\
Top     & 0.17, 0.07 & 0.26, 0.09 \\
Trousers & 0.11, 0.04 & 0.17, 0.05 \\
Skirts  & 0.12, 0.05 & 0.18, 0.06 \\
\midrule
\multicolumn{3}{l}{\textbf{VR\_folding (Second Folds)}} \\
Tshirts & 0.20, 0.10 & 0.29, 0.12 \\
Top     & 0.25, 0.10 & 0.33, 0.11 \\
Trousers & 0.22, 0.08 & 0.29, 0.09 \\
Skirts  & 0.24, 0.08 & 0.33, 0.10 \\
\bottomrule
\end{tabular}
\vspace{-10pt}
\end{table}
\section{Evaluations}
We evaluate how well the dGLI disk and hence the CloSE representation captures the fold location, We test this on 3 datasets: \looseness=-1
\begin{enumerate}
    \item \textbf{The Clothilde dataset}: Dataset we generated with MATLAB using the Clothilde simulator presented in \cite{COLTRARO2024}, with 46 samples
    \item \textbf{The Napkin dataset} from \cite{Borras2023virtual} up to first fold, with 204 samples
    \item\textbf{The VR Folding dataset} from \cite{xue2023garmenttracking} with 5336 samples.
\end{enumerate}
All the datasets contain the start and the end configuration of folding sequences. The Clothidle dataset has few but diverse examples, i.e., clothes of different shapes (Napkins, T-shirts, Circular towels and Trousers) but planar, that is, like towels in different shapes. The napkin dataset has more examples of different folds on a napkin. Though this dataset contains only napkins, the dataset is generated in a virtual environment by tracking human actions on the cloth, thereby being more realistic and has a less accurate mesh and more noise. This dataset also contains $\approx$40\% examples where the cloth is placed at varied location and orientation. 
The VR Folding dataset, contains simulated but real-like clothes from the CLOTH3D dataset \cite{bertiche2020cloth3d}, unlike the previous two which assume the cloth is a 2D deformable object. This dataset is also generated in virtual environment by a human thus making this dataset more complex than the napkin dataset, real-like and very noisy. Also, this dataset is considerably bigger than the other two, containing 3083 first fold examples and 2253 second fold examples. We used the method proposed in \cite{bifold} to discretize the continuous VR folding dataset.

\begin{figure}[tb]
    \centering
    \includegraphics[width=0.85\linewidth]{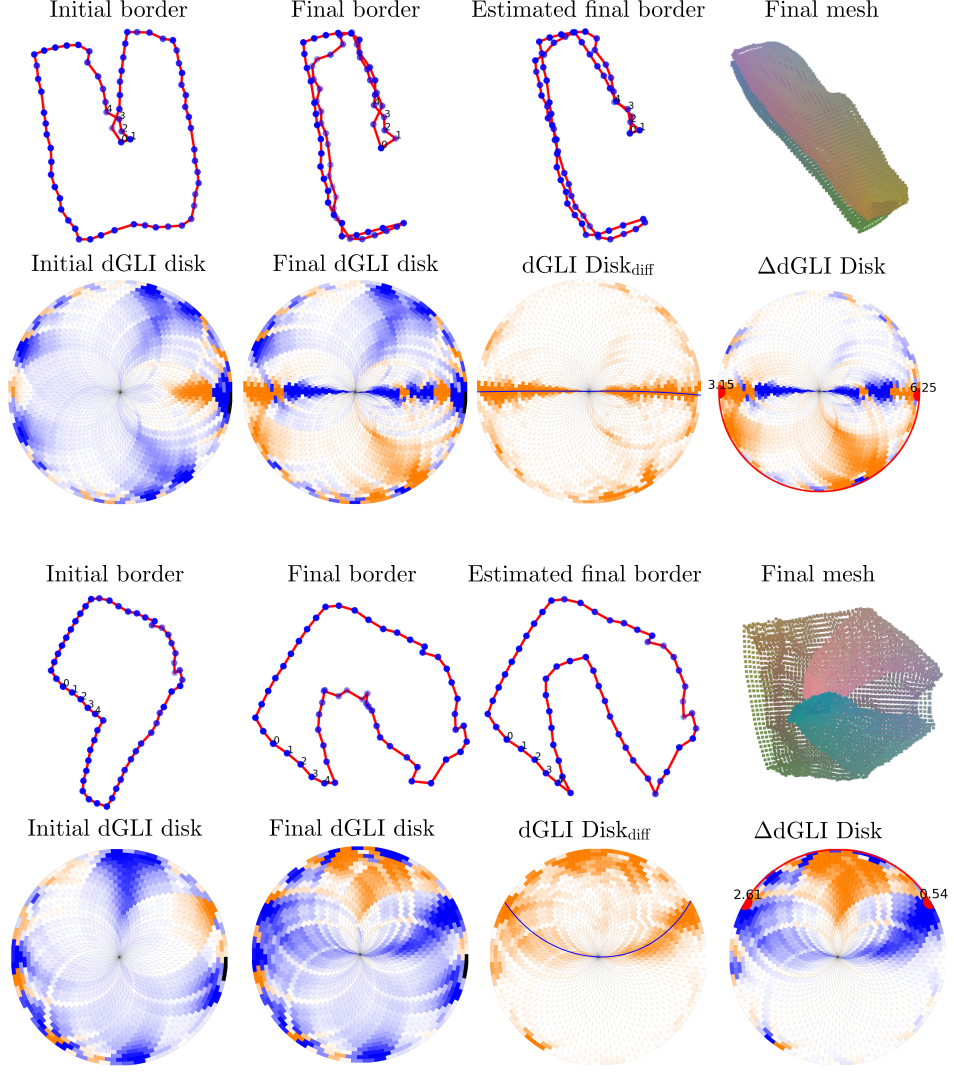}
    \caption{Examples for the VR folding dataset: the first fold (top) and the second fold (bottom).\\
    The second rows show the dGLI disks, start and end, along with the \dGLIdiff\ and the \dGLIdiffsimp\ indicating the fold curve and side folded respectively}
    \label{fig:samples}
    \vspace{-10pt}
\end{figure}

To retrieve the fold location from the dGLI Disk, we follow the optimization described in Section \ref{CloSE}. We set the hyper parameters $k_1, k_2, k_3$ to 0.1, 0.05 and 0.1 respectively while $\tau$ was set to $\pi/4$. For every sample, we estimate the location of the folds. With that information, we can estimate the border after the fold by flipping the side that folds from the initial cloth border along the estimated fold location. We then compare the  actual final cloth border with the estimated final cloth border by calculating the Root Mean Squared Error (RMSE) and the Frechet distance between the two curves. Results are shown in Table \ref{tab:eval}.
Fig. \ref{fig:samples} shows two examples of the initial and final borders, the estimated final border and the corresponding dGLI Disk representations. 
For fair comparison between different shapes, all examples are scaled to exactly fit into a unit circle with the centroid of the border being the center of the circle. 

Unlike the first two datasets, the \textit{VR Folding dataset} provides a full cloth mesh that is closer to real garments. In this case, the mesh border does not directly correspond to the silhouette border we need, requiring a method to estimate it from the start configuration. To do so, we first take the depth image of the cloth in the zenithal direction, mask out the cloth and extract the silhouette border from this mask. We then retrace this border back onto the mesh. This process yields more points than desired leading to a border curve with branches. 
To obtain a cleaner closed single curve, we construct an $\epsilon$-cover\footnote{An $\epsilon$-cover is a set of balls of radius $\epsilon$ that collectively cover all points in the set} on the set of all border vertices %
and use only the centers of these $\epsilon$-balls as the cloth border vertices. Once this border is obtained, the remainder of the procedure follows the same steps as for the first two datasets. For our experiments we used $\epsilon = 0.04$

\begin{figure}[tb]
    \centering
    \resizebox{0.85\linewidth}{!}{%
    \begin{tabular}{ccc}
        \textbf{Actual} & \textbf{Predicted} & \textbf{Fitted Curve} \\[-0.3em]
        \includegraphics[width=0.20\textwidth]{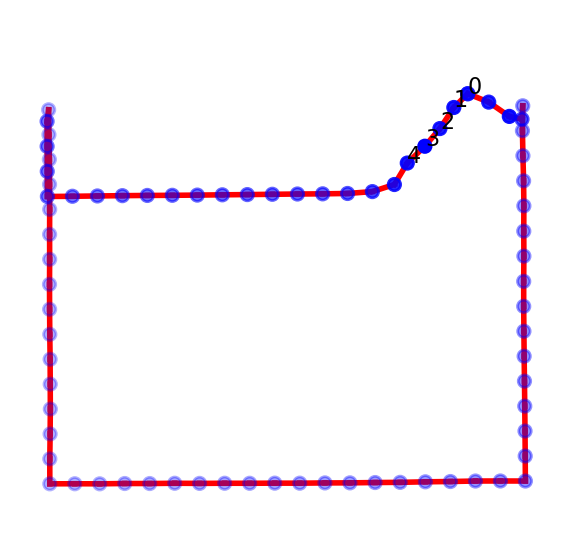} &
        \includegraphics[width=0.20\textwidth]{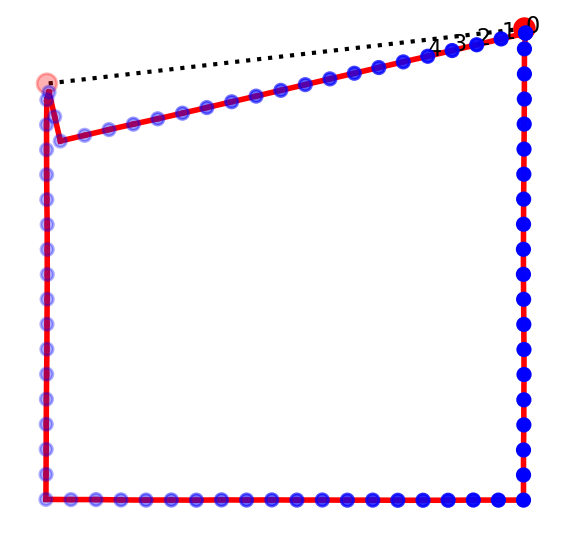} &
        \includegraphics[width=0.25\textwidth]{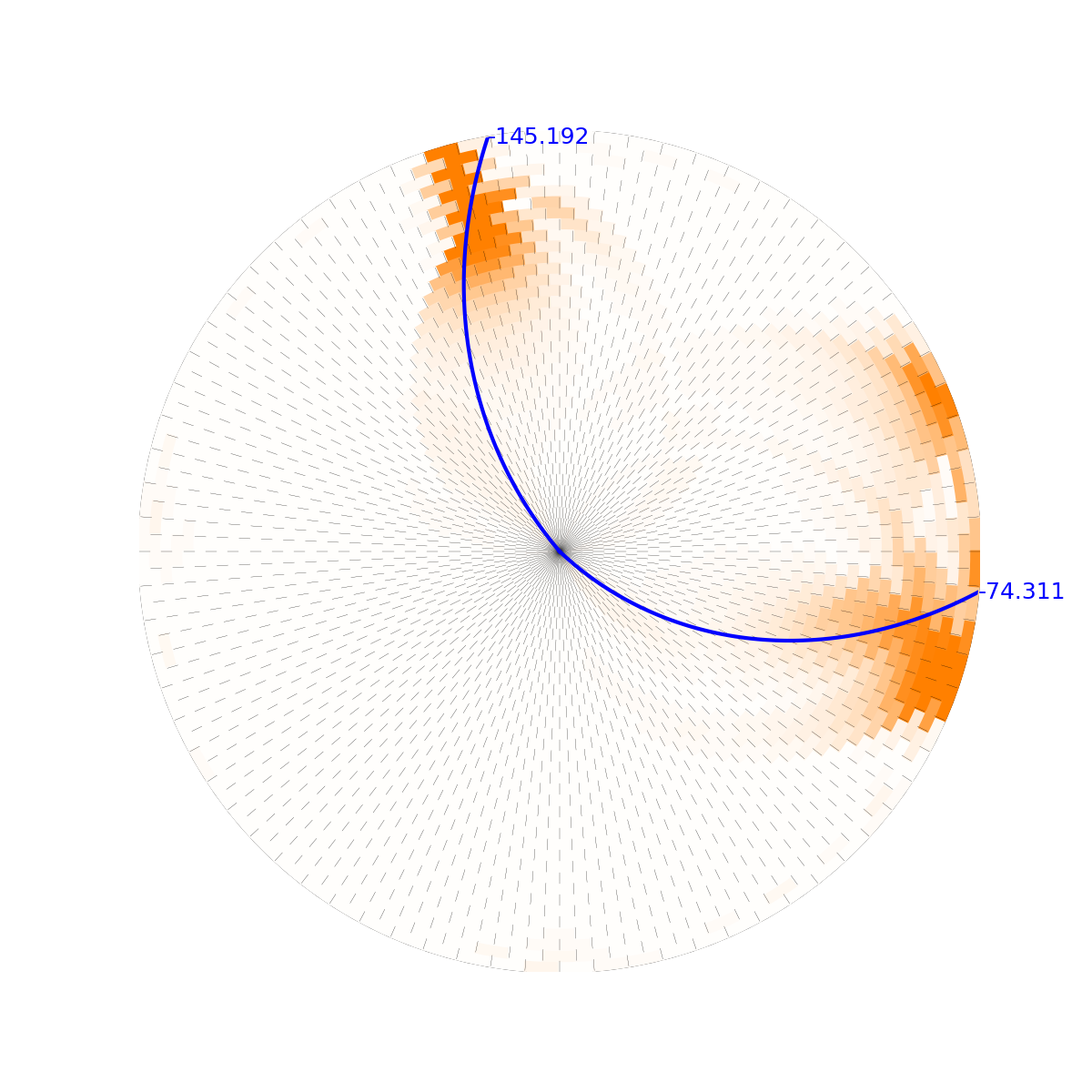} \\
        \multicolumn{3}{c}{\#87 (Clothilde dataset) -- Inaccuracy due to unclean fold.} \\[-0.3em]
        \includegraphics[width=0.20\textwidth]{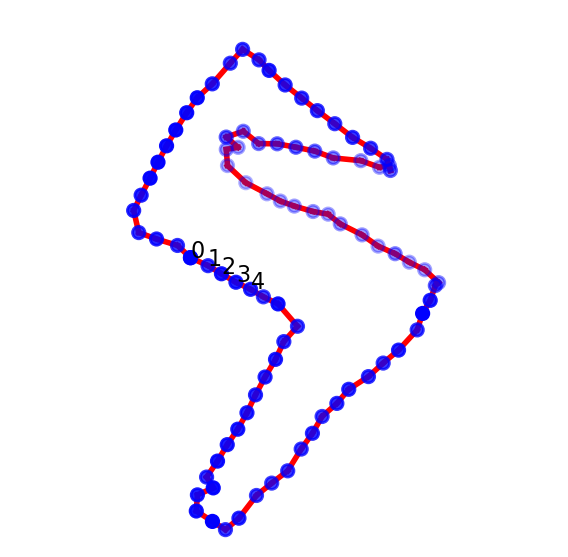} &
        \includegraphics[width=0.20\textwidth]{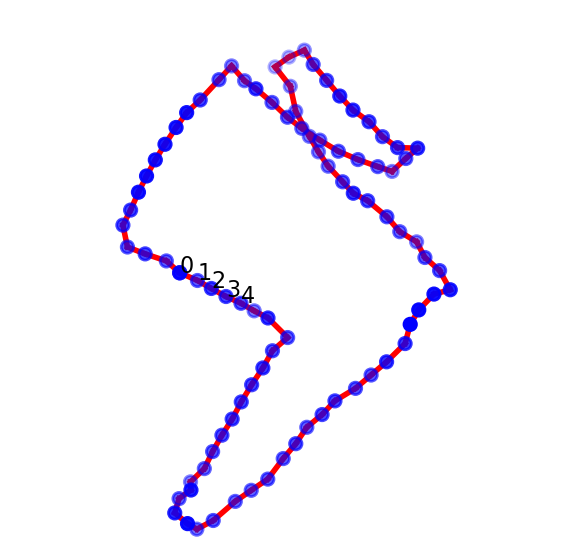} &
        \includegraphics[width=0.25\textwidth]{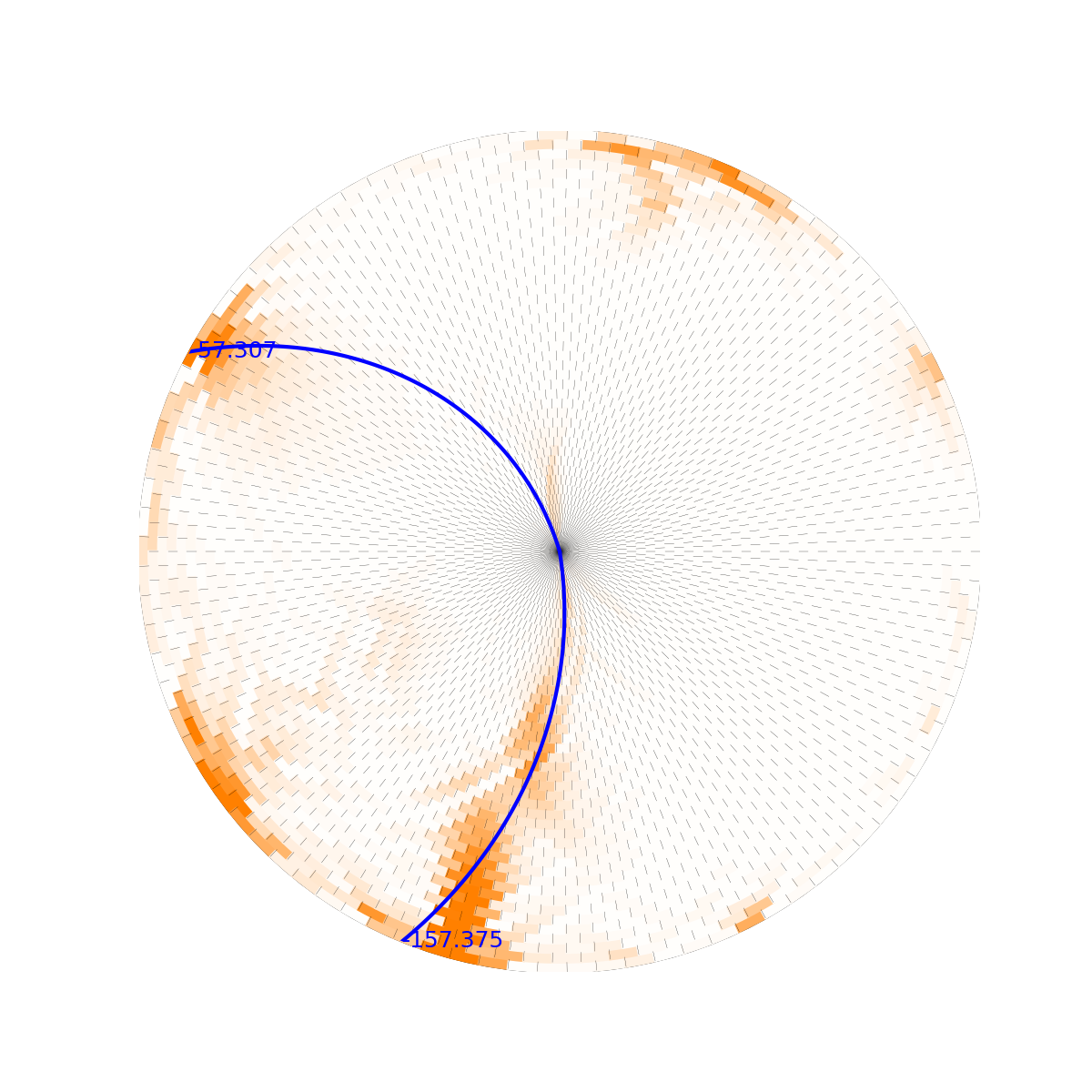} \\
        \multicolumn{3}{c}{Tshirt from VR folding -- the shape of the cloth amplifies distance reading.} \\
    \end{tabular}}
    \caption{Examples of ground truth borders, predicted borders, and fitted curves that give the highest errors.}
    \label{fig:highError}
    \vspace{-10pt}
\end{figure}

\begin{figure*}[bt]
    \centering
    \includegraphics[width=0.85\linewidth]{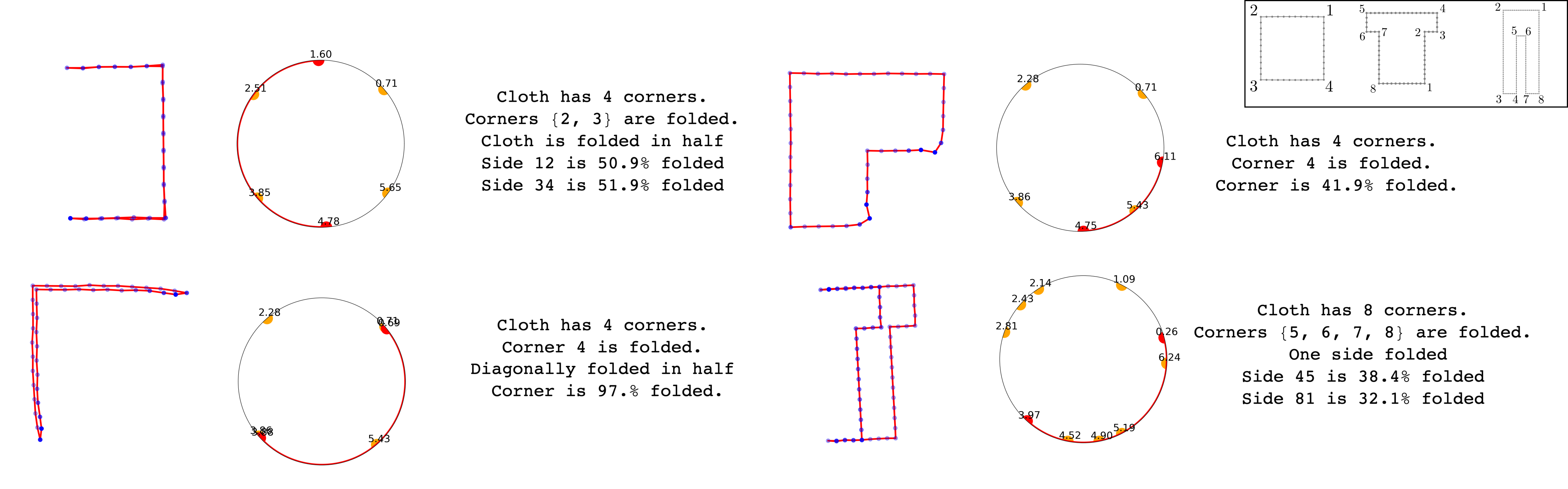}
    \caption{Different examples of the CloSE\ representation and the corresponding automatic generated label. More examples can be found in the paper website.}
    \label{fig:semanticLabelling}
    \vspace{-10pt}
\end{figure*}

\subsection{Analysis}
Table. \ref{tab:eval} shows how we consistently get good results in all the three datasets. The Clothilde dataset is the cleanest hence we see the lowest error and the prediction error increases as the data becomes more noisy. Note that these predictions are still very accurate even for the noisiest, VR folding dataset (see examples in Fig. \ref{fig:samples}, and many more on the paper website). \looseness=-1

Fig. \ref{fig:highError} shows the ground truth border and the predicted border for the worst-case examples. Biger errors are majorly due to two reasons, First, if the fold is improper (Fig.\ref{fig:highError}[top]), i.e., there is an extra fold in addition to the primary fold confusing the dGLI disk. Second, if the shape is highly concave and the fold occurs in the concave section (Fig.\ref{fig:highError}[bottom]). Here a slight error in fold location prediction leads to a big difference in the final meshes creating a high rmse loss. This kind of error does not indicate any fault in our algorithm but is rather a result of strictness of the measure we used. \looseness=-1

All the examples in the Clothilde and the Napkin datasets predicted the existence of exactly one fold curve. However, the VR Folding dataset exhibited errors in $4\%$ and $10\%$ of the cases for the first and second folds, respectively. These errors were not due to limitations of our representation, but rather because the method to estimate the initial border curve failed to find the correct border or because the fold was very bad causing our method to detect more than one fold.

\section{Applications}
We present two applications that arise naturally from our CloSE representation: automatic semantic labeling and high- and low-level planning. %

\subsection{Semantic Labeling}
Given the CloSE representation of a cloth state 
\begin{equation}
    ( (\theta_1, \dots, \theta_n), (f_1, f_2)),
    \nonumber
\end{equation}
where $n$ is the number of corners, by construction each $\theta_i$ and $f_i$ are numbers from  $[0, 2\pi)$, because they are angles in radians.  Note that $\theta_i$ are always in ascending order and correspond to the corners of the cloth in order following the border curve. That is, given the border as a sorted list of vertices, $\theta_1$ is the coordinate corresponding to the first corner we find in that list. Some examples of the used order are shown at the top right of Fig. \ref{fig:semanticLabelling}.

Therefore, we can identify what is the semantic state of cloth by using interval reasoning in the circle. Each fold is an interval and we can identify which  of the $\theta_i$ are inside, and these are the folded corners. In addition, each $f_i$ falls inside an edge of the cloth, that is 
\begin{equation}
    \theta_j < f_i < \theta_{j+1} \text{ for any } i\in {1,2}, j\in {1\dots n},
    \nonumber
\end{equation}
accounting for the $2\pi-0$ equivalence. The distances of $f_i$ to the neighbors $\theta_i, \theta_{i+1}$ are proportional to the distance of where the fold is on the folded edge of the corresponding corner. Therefore, we can also identify where the fold is on the folded edge. 

By simple reasoning we can identify if the cloth is folded in a symmetric half, if the two folded edges are both folded by the same proportion, etc. We cannot identify the shape but we can say how many corners it has. Several examples of the obtained labels are showed in Fig. \ref{fig:semanticLabelling}. Note that we obtain these without any learning, by just reasoning on our feature vector, and depending on the desired type of labels we can give more abstract or more informed labels. You can find the code to generate the labels in the accompanying website.

\begin{figure}[b]
    \begin{subfigure}[b]{\linewidth} %
        \centering
        \begin{adjustbox}{valign=c}
            \includegraphics[width=0.5\textwidth]{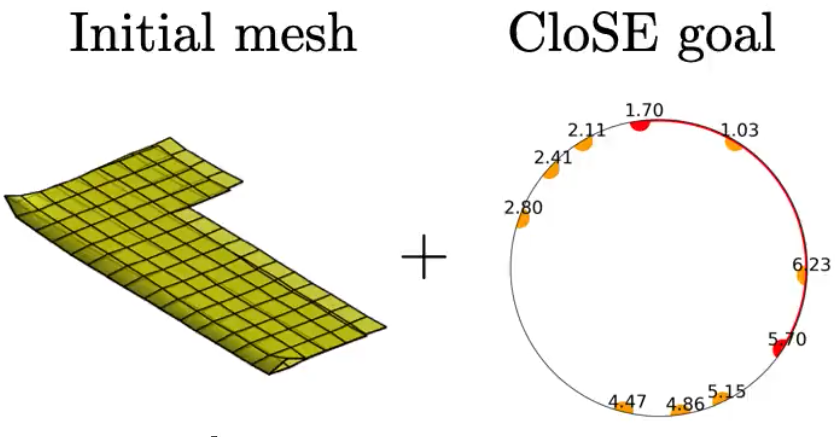}
        \end{adjustbox}
        \hfill
        \begin{adjustbox}{valign=c}
            \includegraphics[width=0.45\textwidth]{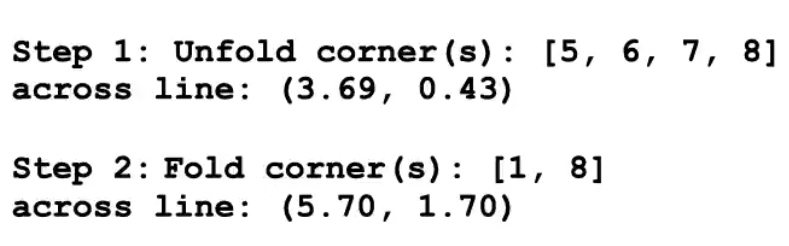}
        \end{adjustbox}
        \caption{Inputs (Left) and Output (Right) for the planning Case 2}
        \label{fig:planning_viaBase}
    \end{subfigure}
    \begin{subfigure}[b]{\linewidth} %
        \centering
        \includegraphics[width=0.2\textwidth]{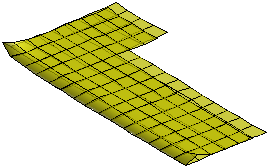}
        \includegraphics[width=0.24\textwidth]{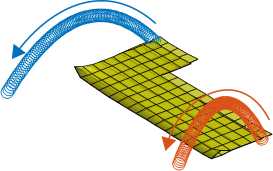} %
        \includegraphics[width=0.24\textwidth]{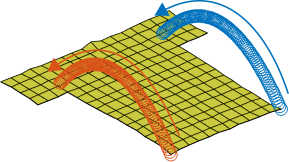} %
        \includegraphics[width=0.2\textwidth]{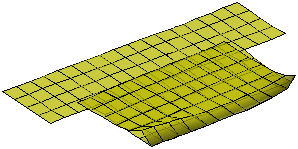}
        \caption{Fold Visualization - Case 2}
        \label{fig:multi-stepFoldPrediction}
    \end{subfigure}
    \caption{Planning (Case 2): Goes through unfolded intermediate state}
    \label{fig:planning}
    \vspace{-10pt}
\end{figure}

\subsection{Planning}
Given the initial border of the cloth and the CloSE representation of the desired fold state, our system can automatically plan a sequence of manipulations, providing both semantic (high-level) instructions and low-level trajectories for the corners to be manipulated. \looseness=-1

Because the CloSE representation explicitly tracks the fold locations and categorizes them into semantic regions, we devise a rule-based algorithm to reason over the cloth's current state and plan manipulation actions. Fig. \ref{fig:planning} shows one of the two cases that we encounter when we are dealing with one fold.\footnote{The two cases can be seen on the website. We also run the trajectories in simulation, and these can be seen in the video.}

In Case 1, both the initial and goal configurations lie in the same semantic region (i.e., the fold encloses the same corners). In this scenario, the algorithm determines that no intermediate states are needed and directly plans a single-step manipulation to move the currently folded corner(s) to their goal position(s). 

In Case 2, the initial and goal configurations belong to different semantic states. To resolve this, our algorithm employs a multi-step approach where the cloth must pass through the unfolded (intermediate) state. Thus, the planner deterministically generates an "unfold" action followed by the new "fold" action.

For low-level planning, i.e., deciding the exact pick-and-place points, the CloSE representation provides the necessary geometric features. When a fold requires moving more than two corners, our system automatically selects the optimal pair of corners, i.e., corners that maximize the area of the trapezoid formed between those corners and the fold line. The underlying idea is to maximize the area of the cloth controlled by the robot during the maneuver. Our generated manipulation actions are shown in Fig. \ref{fig:multi-stepFoldPrediction}.

\section{Multi-folds (Future Work)}\label{sec:multi-folds}
\begin{figure}[b]
    \vspace{-10pt}
        \centering
        \includegraphics[width=\linewidth]{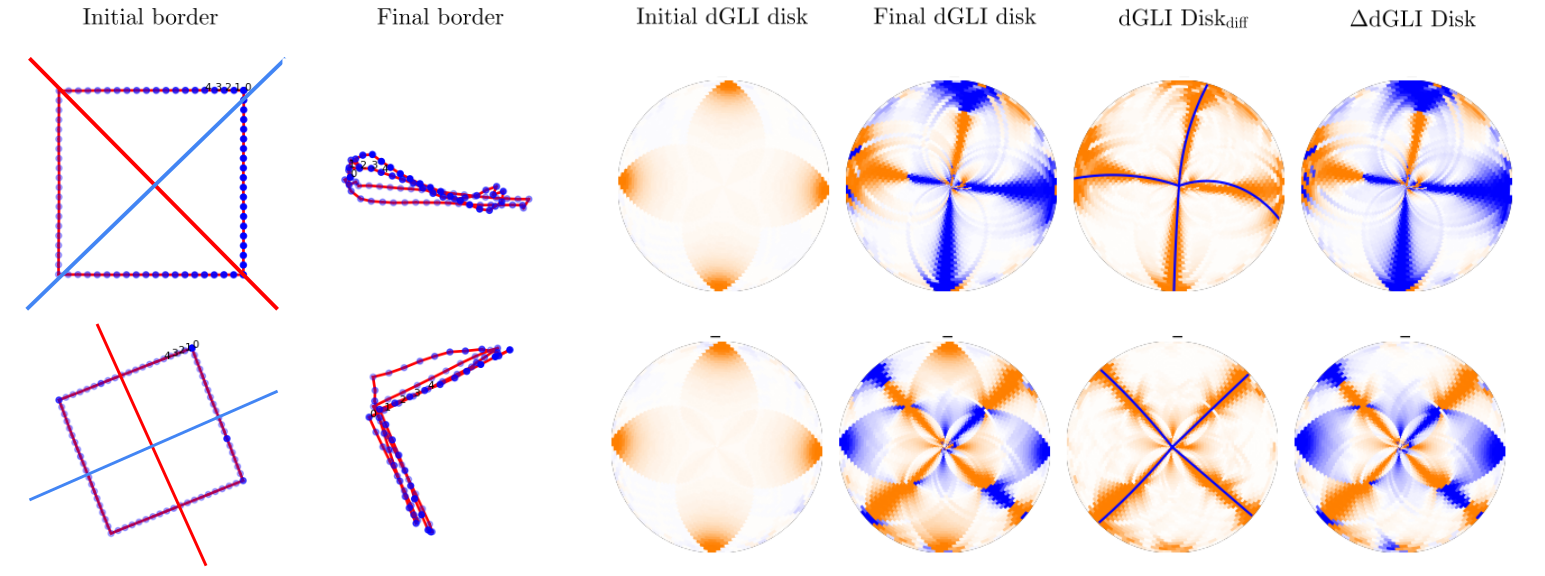}
        \caption{
        Multi-fold patterns in the dGLI disk. Colored lines on the cloth border indicate distinct folds. While the disk captures fold locations, interpreting the intricate features of simultaneous folds remains future work.
        }
        \label{fig:multifold}
    \vspace{-10pt}
    \end{figure}

    While our discussion has primarily focused on single folds, the dGLI disk also captures compelling patterns in the presence of multiple folds. In Fig. \ref{fig:multifold} we show the results of applying our method to estimate the folds of a napkin after applying two diagonal folds (top example) or two half folds (bottom). From the dGLI disk of the starting and final states we compute the dGLI diff and apply our optimization to estimate the fold lines that appear drawn in the figure. We can find four lines that are associated with the two folds. This shows promising results. However, when the sequence of folds is more complex, this correspondence becomes more intricate and we leave it for future work.

    Importantly, our current framework is already capable of managing multi-fold scenarios through sequential, iterative applications of single folds. Our evaluations demonstrate that the dGLI disk successfully and robustly detects subsequent fold lines, even for the highly noisy VR folding dataset examples (see Fig.\ref{fig:samples}(bottom)).

\section{Conclusion}
    In summary, we discuss a novel geometric representation of clothes states. First, we introduce the dGLI disk, an intermediate representation that is a rearrangement of the dGLI coordinates onto a disk. This representation reveals easy-to-detect geometric features that characterize the cloth folding state independently of the cloth shape. The dGLI disk representation could be directly fed as an extra feature of a learning method, providing generality to shape just by construction.
    
    We then abstract the features from our dGLI disk onto a circle. This circular representation, CloSE, is compact and continuous while maintaining the generality for different cloth shapes offered by the dGLI disk. 
    We then empirically show that the fold locations are accurately captured by our geometric representation, which, to our knowledge, is an unexplored topic even in simulation. 
    Finally, we show two important applications that come naturally from this representation: Semantic labeling and high- and low-level fold planning. \looseness=-1

    A good cloth representation is the one that allows a robot to reason about cloth states, and plan a sequence of motions to attain a target cloth configuration. Devising a powerful representation for clothes is a very important and difficult problem in robotics and computer graphics. This work is a step in that direction. \looseness=-1

\bibliographystyle{ieeetr}
{
\balance
\small
\bibliography{main}
}

\end{document}